\documentclass{article} % For LaTeX2e
\usepackage{graphics}
\title{Bias Plus Variance Decomposition for Survival Analysis Problems}

\author{
Marina Sapir \\
metaPattern\\
Bar Harbor, ME 04609 \\
\texttt{m@sapir.us} \\
}

\begin{document}

\maketitle

\begin{abstract}
 Bias - variance decomposition of the expected error defined for regression and classification problems is an important tool to study and compare different algorithms, to find the best areas for their application. Here the decomposition is introduced for the survival analysis problem. In our experiments, we study bias -variance parts of the expected error for two algorithms: original Cox proportional hazard regression and  CoxPath, path algorithm for $L_1$-regularized Cox regression, on the series of increased training sets.  The experiments demonstrate that, contrary expectations, CoxPath does not necessarily have an advantage over Cox regression. 
 \end{abstract}

\section{Introduction}

For classification problems, it is well known that bias and variance components of the estimation prediction error combine to influence classification in a very different way, and have different importance depending on the sample size. For small and for high-dimensional datasets, variance of the prediction caused by variations in the training samples makes largest contribution into the expected prediction error. For large datasets, bias becomes more important component of the error \cite{Friedman}.

Thus, the decomposition of expected error into bias and variance parts is an important tool to understand differences between the algorithms, to find areas of the optimal application.

To the best of author's knowledge, such decomposition was not proposed for survival analysis problem. Here we describe an approach to define this decomposition for this class of problem. On two real life datasets we study how bias and variance of we show how regularization and  size of the training sample affect bias, variance and overall errors of the methods.

\section{Bias - Variance Decomposition}

\subsection{Survival analysis problem}
Survival analysis  deals with the  datasets, where each observation has three components:   covariate vector $x$, a positive survival time $t$ and an event indicator $\delta$, which  is equal to 1 if an event (failure) occurred, and zero if the observation is (right) censored at time $t.$ 

The prediction in survival analysis is generally understood as an estimate of an individual's risk, but the concept of the risk is open for interpretation. The commonly accepted  criterion of the accuracy of the risk modeling is Harrell's concordance index \cite{Tutorial} measuring agreement between the model's scores and the order of the failure times. The criterion is not directly related with any particular interpretation of the scores. 

Because of the presence of censored observations, failure times define only partial order on observations. Two observations $c_1 =\{x_1, t_1, \delta_1\}$ and $c_2 = \{x_2, t_2, \delta_2\}$ are ordered $c_1 \prec c_2$ if and only if $t_1 < t_2$ and $\delta_1 = 1$. In case of absent ties, the concordance index equals proportion of correctly ordered (concordant) pairs of observations: 
$$CI = \frac{concordant +0.5 \cdot ties}{discordant + concordant + ties}.$$ 

Then the survival analysis can be considered as a problem discerning between concordant and discordant pairs of observations.   If the features are continuous, the ties are rare, and  proportion of the concordant pairs closely approximates concordance index.   This allows us to study bias-variance decomposition for survival analysis using available bias -variance decomposition for the classification problems.

\subsection{Bias - variance decomposition for classification problems}

For the binary classification problems, the commonly used bias-variance decomposition of the classification error $E(C)$ is proposed in \cite{Kohavi}:
\begin{eqnarray*}
E(C)         & = &  0.5 \cdot bias^2(x) + 0.5 \cdot variance^2(x) + 0.5 \cdot \sigma^2(x), \\
bias^2_x  & =  &  \frac{1}{2}\sum_{y \in \{0,1\} } \left[P(Y_F = y | x) - P(Y_H = y | x) \right]^2\\
variance _x & = & \frac{1}{2} \left(1 - \sum_{y \in \{0,1\} } P(Y_H = y | x)^2 \right) \\
\sigma_x^2 & = & \frac{1}{2} \left( 1 - \sum_{y \in \{0,1\}} P(Y_F = y | x)^2 \right), 
\end{eqnarray*}
where $Y_H$ is the classification obtained on the training set $H$, $Y_F$ is the actual class values. 
The bias is a measure of closeness between the distributions of values  $Y_H(x)$ over the training sets $H$ of a fixed size, and the distribution of $Y_F(x),$ $\sigma_x^2$ represents the level of noise in the class variable, and the $variance_x$ is an estimate of variability of the decisions on the training sets of the given size.   

The more data - sensitive is the learning algorithm  the less bias it has. The notion of bias - variance tradeoff  \cite{Friedman} refers to the fact that the lower is the algorithm's bias, the larger shall be dependence of the learned function on the training set, especially when number of the training cases is small, or the dimensionality of the data is high. 

Bias and variance of algorithms depend on the size of the training set, complexity of the learned function, and many other factors, including specifics of  particular data.  Here we explore these components for two algorithms on two real datasets. 
 
 \section{The algorithms under comparison}
 
In the traditional approach associated with sir David Cox \cite{Cox}, the research is concentrated on a time-dependent  ``hazard function''  $\Lambda(x, t)$:  event rate at time $t$ conditional on survival of the individual $x$ until time $t$ or later (that is, $T \ge t$). 

 Cox proportional hazard (PH) regression is based on the strong assumption that the hazard function has the form of $$\Lambda(x, t) = \lambda(t) \cdot  exp(\beta(x)),$$ where $\lambda(t)$ is  unknown time-dependent function, common for all individuals in the population. The assumption implies, in particularly, that for any two individuals, their hazards are proportional all the time.  
So, the result of the modeling is, actually,  not the individual time-dependent hazard functions, but rather these ``proportionality'' scores. 

Most of  advanced methods for prediction in survival analysis are developed to make this traditional approach more robust against overfitting on sparse data (see surveys in \cite{Segal, Survey2}). Among the regularization methods, $L_1$ -penalized Cox regression is the most attractive because it produces concise interpretable rules. The CoxPath \cite{Park} is path algorithm which builds $L_1$ regularized proportional hazard regression models with series of values of the regularization parameter $\lambda$, and then it selects one of the solution based on the  performance criterion. Regularization lowers algorithm variance, making an individual regression model more robust against variations between  small training sets.  Selection of the best model is intended to improve the bias of the algorithm. According to the bias - variance tradeoff, neither step necessarily decreases overall prediction error.  In the next section we describe the results of the experiments   evaluating the bias and variance of these algorithms on series of increased training sets. 

\section{Computational Experiments}

For real life datasets, only variance and prediction error can be measured directly. The sum of bias and measure of noise $\sigma^2$ constituting  unavoidable error was evaluated as the difference between the prediction error and the variance. 

In the experiments, first, 20$\%$ of the whole sample was set aside as a test set. From the rest, increasing subsets of the data were randomly selected as training sets; 20 training sets of each size were selected. All the methods were trained on the training sets and the models applied on the single test set to evaluate variance on the test data.  The procedure was repeated 10 times with randomly chosen test sets, and average variance and performance for each training set size was evaluated across all 10 test sets.

The experiment was conducted on two datasets. 
\begin{itemize}
\item PBC : This data is from the Mayo Clinic trial in primary biliary cirrhosis of the liver conducted between 1974 and 1984 \cite{PBC}. Patients are characterized by standard description of the disease conditions.   The dataset has 17 features and 228 observations.

\item Ro02s: the dataset from \cite{Ro02} contains information about 240 patients with lymphoma. Using hierarchical cluster analysis on whole dataset and expert knowledge about factors associated with disease progression, the authors identified relevant four clusters and a single gene out of the 7399  genes on the lymphochip. Along with gene expressions, the data include two features for histological grouping of the patients. The authors aggregated gene expressions in each selected cluster to create a ÒsignaturesÓ of the clusters. The signatures, rather than gene expressions themselves were used for modeling.  The dataset with aggregated data has 7 features. 
\end{itemize}

\begin{figure}[h]
\begin{center}
%\framebox[4.0in]{$\;$}
%\fbox{\rule[-.5cm]{0cm}{4cm} \rule[-.5cm]{4cm}{0cm}}
\includegraphics{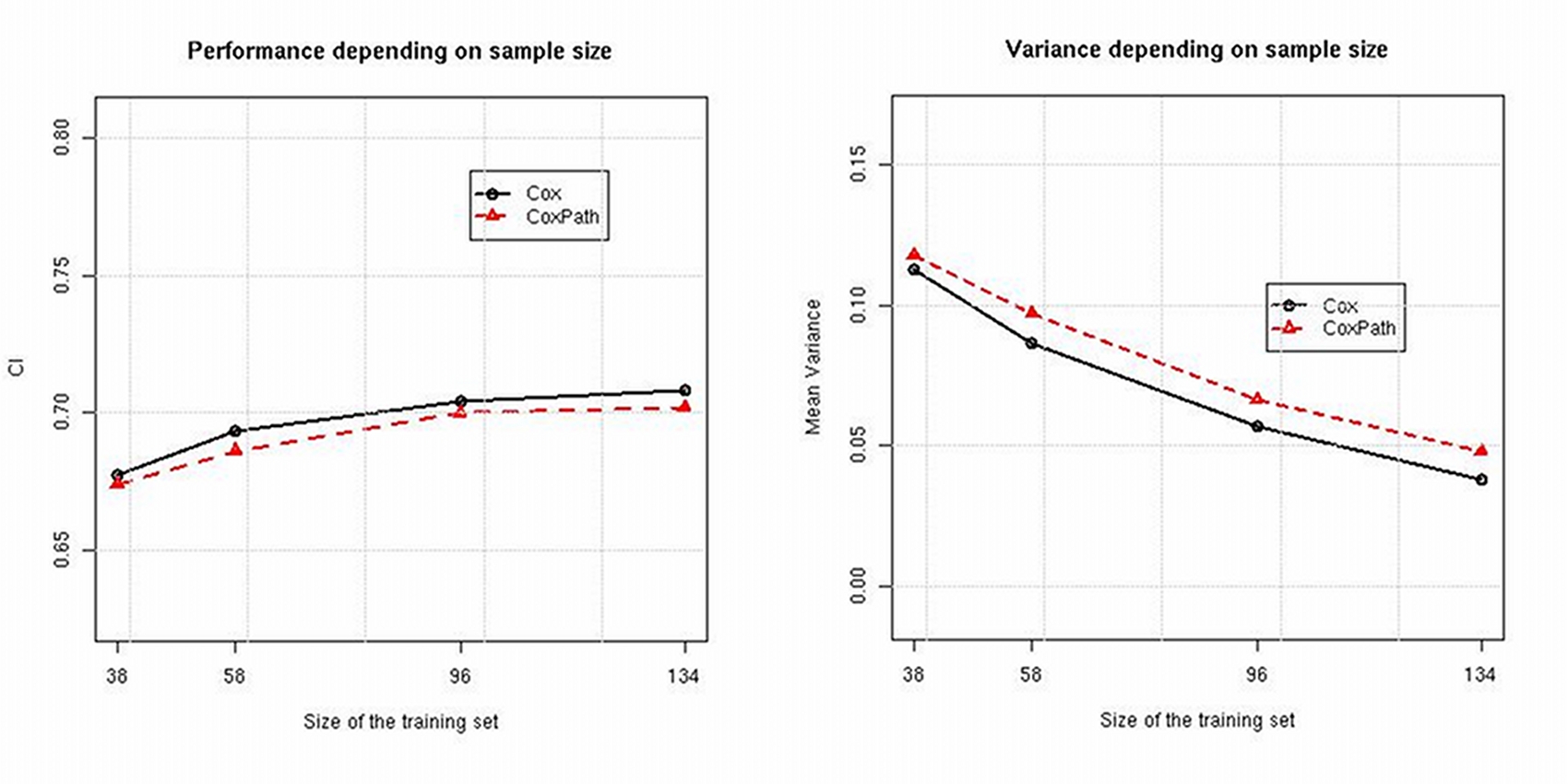}
\end{center}
\caption{Ro02 dataset, Variance and Performance of Cox PHR and CoxPath}
\end{figure}

\begin{figure}[h]
\begin{center}
%\framebox[4.0in]{$\;$}
%\fbox{\rule[-.5cm]{0cm}{4cm} \rule[-.5cm]{4cm}{0cm}}
\includegraphics{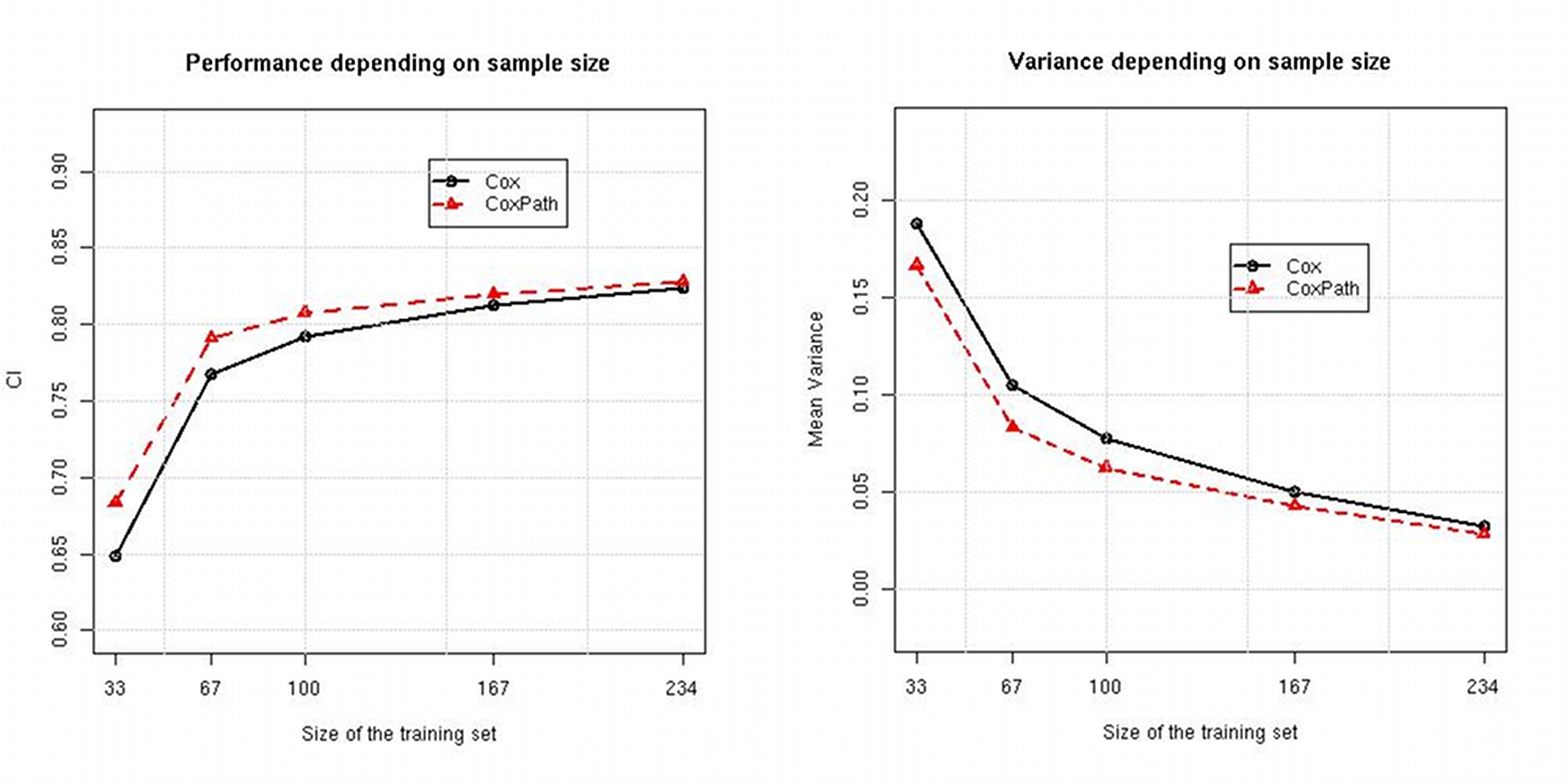}
\end{center}
\caption{PBC dataset, Variance and Performance of Cox PHR and CoxPath}
\end{figure}

The results are presented on the Figures 1, 2, where we show methods performance as $1 - E(C).$ Bias of both methods is almost indistinguishable on both datasets, and is not shown here.  The figures show that CoxPath does not have consistent advantage over Cox PH regression. On PBC dataset, CoxPath has lower variance and better performance for all sample sizes, while for Ro2 dataset the opposite is true. 

One can hypothesize  that an advantage in variance CoxPath obtained due to regularization was offset by the additional sensitivity to the training data due to the model selection.

Additional experiments with artificial datasets and $L_1$-regularized Cox PH regression with fixed parameter $\lambda$ may help better understand the factors affecting bias and variance of the methods and to produce recommendations for the types of data, for which one or another method is preferable. 
\small{

\bibliography{survival}
}
\end{document}